\documentclass[]{memtensor}

\usepackage{enumitem}
\usepackage[toc,page,header]{appendix}
\usepackage[utf8]{inputenc} % allow utf-8 input
\usepackage[T1]{fontenc}    % use 8-bit T1 fonts
\usepackage{hyperref}       % hyperlinks
\usepackage{url}            % simple URL typesetting
\usepackage{booktabs}       % professional-quality tables
\usepackage{amsfonts}       % blackboard math symbols
\usepackage{nicefrac}       % compact symbols for 1/2, etc.
\usepackage{microtype}      % microtypography
\usepackage{amsmath} 
\usepackage{etoolbox}
\usepackage{lipsum}  
\usepackage{minitoc}
\usepackage{tablefootnote}  % footnote in table
\usepackage{threeparttable}
\usepackage{wrapfig}
\title{ \includegraphics[width=0.04\linewidth]{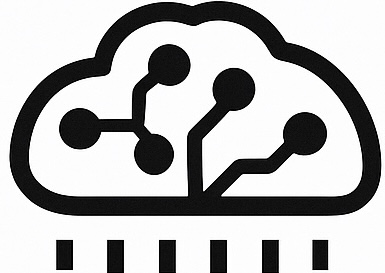}\hspace{1mm}MemOS: An Operating System for Memory-Augmented Generation (MAG) in Large Language Models (Short Version)}

\author[*]{Zhiyu Li}
\author[1,3,*]{Shichao Song}
\author[1,3,*]{Hanyu Wang}
\author[3,*]{Simin Niu}
\author[4,*]{Ding Chen}
\author[1,3]{Jiawei Yang}
\author[1]{Chenyang Xi}
\author[3]{Huayi Lai}
\author[3]{Jihao Zhao}
\author[1]{Yezhaohui Wang}
\author[1]{Junpeng Ren}
\author[1]{Zehao Lin}
\author[1]{Jiahao Huo}
\author[2]{Tianyi Chen}
\author[1]{Kai Chen}
\author[2]{Kehang Li}
\author[3]{Zhiqiang Yin}
\author[1]{Qingchen Yu}
\author[1,\dagger]{Bo Tang}
\author[1,\dagger]{Hongkang Yang}
\author[2,\dagger]{Zhi-Qin John Xu}
\author[1,\dagger]{Feiyu Xiong}

\affiliation[1]{MemTensor (Shanghai) Technology Co., Ltd.}
\affiliation[2]{Shanghai Jiao Tong University}
\affiliation[3]{Renmin University of China}
\affiliation[4]{Research Institute of China Telecom}

\abstract{
Large Language Models (LLMs) have emerged as foundational infrastructure in the pursuit of Artificial General Intelligence (AGI). Despite their remarkable capabilities in language perception and generation, current LLMs fundamentally lack a unified and structured architecture for handling memory. They primarily rely on parametric memory (knowledge encoded in model weights) and ephemeral activation memory (context-limited runtime states). While emerging methods like Retrieval-Augmented Generation (RAG) incorporate plaintext memory, they lack lifecycle management and multi-modal integration, limiting their capacity for long-term knowledge evolution. To address this, we introduce —a memory operating system designed for LLMs that, for the first time, elevates memory to a first-class operational resource. It builds unified mechanisms for representation, organization, and governance across three core memory types: parametric, activation, and plaintext. At its core is the \textbf{MemCube}, a standardized memory abstraction that enables tracking, fusion, and migration of heterogeneous memory, while offering structured, traceable access across tasks and contexts. \textsc{MemOS} establishes a memory-centric execution framework with strong controllability, adaptability, and evolvability. It fills a critical gap in current LLM infrastructure and lays the groundwork for continual adaptation, personalized intelligence, and cross-platform coordination in next-generation intelligent systems.
}
\date{2025.05.27}
\correspondence{\email{\{tangb,yanghk,xuzq,xiongfy\}@memtensor.cn}}
\checkdata[Author Legend]{*Co-equal primary author, \dag{}Correspondence}
\begin{document}
\maketitle

\section{Introduction}

Large Language Models (LLMs) are emerging as a foundational pathway toward Artificial General Intelligence (AGI) \cite{zhao2023survey}, yet they remain fundamentally limited in supporting robust memory capabilities. Most current architectures rely on implicit parametric memory—knowledge embedded within massive model weights—which is difficult to interpret \cite{zhao2023explainabilitylargelanguagemodels}, update \cite{meng_memit_2023}, or transfer \cite{hsieh-etal-2023-distilling}. Although Retrieval-Augmented Generation (RAG) incorporates external knowledge sources \cite{DBLP:journals/corr/abs-2402-19473,DBLP:journals/corr/abs-2312-10997,DBLP:journals/corr/abs-2310-05029, DBLP:journals/corr/abs-2404-16130,DBLP:journals/corr/abs-2410-05779}, it effectively serves as an ad hoc textual patch and lacks a structured, unified mechanism for memory management. These architectural shortcomings lead to four critical issues in real-world applications: inability to model long-term and multi-turn conversational states; poor adaptability to evolving knowledge; lack of persistent modeling for user preferences and multi-agent workflows; and the emergence of “memory silos” across platforms, hindering the reuse and migration of prior interactions. At the root of these challenges lies a fundamental oversight: current LLMs do not treat memory as an explicit, schedulable, and governable resource.

To address this, we propose —a memory operating system designed for large language models. \textsc{MemOS} centers memory units as operational resources and establishes a full lifecycle encompassing memory generation, organization, utilization, and evolution. It offers structured representations, unified interfaces, version control, and access governance to overcome systemic limitations in memory handling. Rather than merely extending the RAG paradigm, \textsc{MemOS} introduces a controllable, adaptable, and evolvable memory infrastructure that empowers LLMs to track knowledge updates, internalize user preferences, and maintain behavioral consistency across platforms. This represents a fundamental shift in language model architecture: from systems that merely perceive and generate to those that \textit{remember, adapt, and grow over time}.

\section{Memory in Large Language Models}

\begin{figure}[htp]
    \centering
    \includegraphics[width=1.0\linewidth]{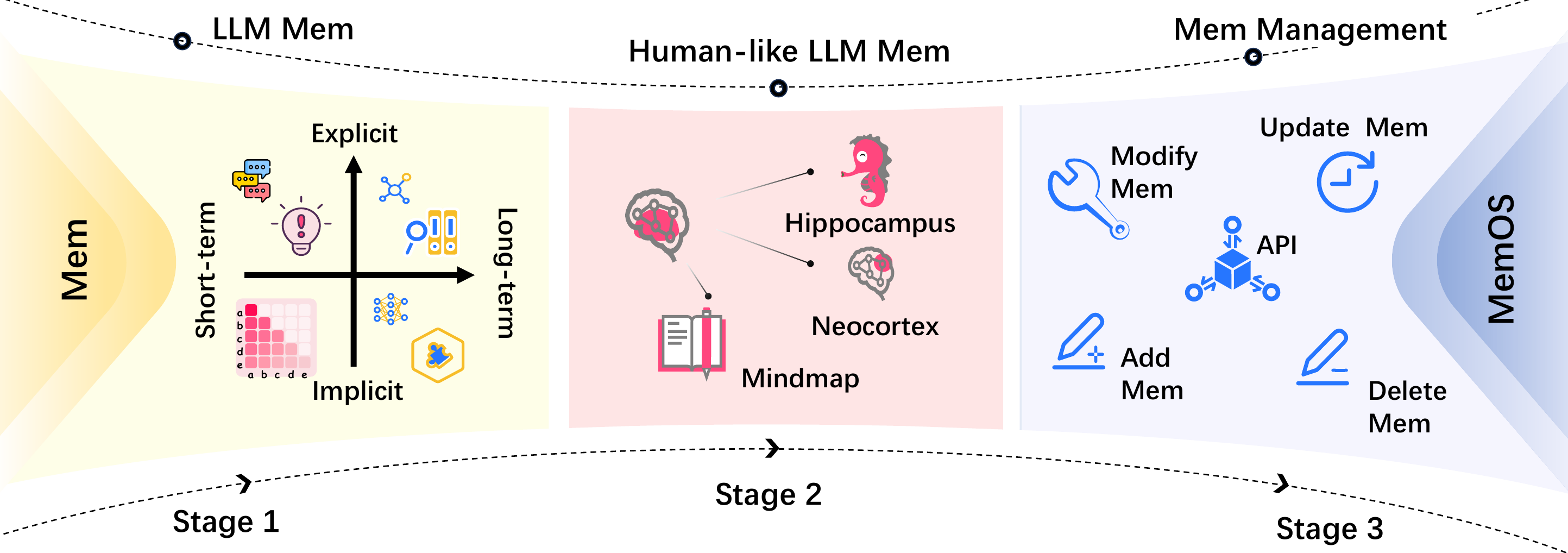}
    \caption{Memory (Mem) in LLMs.} 
    \label{fig:Memory in LLMs}
\end{figure}

Research into LLM memory has progressed through three major stages (see Figure~\ref{fig:Memory in LLMs}).

The first is the \textit{Memory Definition and Exploration} stage, in which researchers classify and analyze memory mechanisms along dimensions such as parametric vs. non-parametric and short-term vs. long-term memory \cite{du_cuhkSurvey_2025, wu_huaweiSurvey_2025, shan_LiAutoSurvey_2025}. For implicit memory, pre-training and adapter-based methods embed knowledge directly into model weights, while knowledge editing techniques enable targeted post hoc modifications \cite{devlin_bert_2019,brown_gpt3_2020,hu_lora_2021,su_PRAG_2025,tan_DyPRAG_2025,meng_rome_2023,cao_KnowledgeEditor_2021,xu_biasedit_2025,fang_alphaedit_2025}. KV-caches and hidden states constitute the core of \textit{implicit short-term memory}, preserving contextual continuity and guiding generation behavior during inference \cite{dong_less_2024, kwon_vLLM_2023,subramani_steerno1_2022, wang_ACT_2025,turner_ActAdd_2024}. Explicit short-term memory typically involves prompt concatenation within the context window, but remains limited by context length constraints \cite{ouyang_instructgpt_2022,liu_lost_2024}. Explicit long-term memory leverages external retrieval mechanisms, increasingly adopting structured formats—such as graphs and trees—to improve semantic integration and retrieval efficiency \cite{khandelwal_knnlm_2019,DBLP:journals/corr/abs-2404-16130,xu2025noderagstructuringgraphbasedrag,yang2025heteragheterogeneousretrievalaugmentedgeneration}.

The second stage involves the \textit{Emergence of Human-like Memory}, where systems optimized for long-term persistence, context awareness, and self-reflection begin to exhibit structural and behavioral patterns reminiscent of human memory. Examples include brain-inspired architectures such as HippoRAG and Memory$^3$ \cite{DBLP:conf/nips/GutierrezS0Y024,yang_memory3_2024}, as well as systems like PGRAG and Second-Me \cite{DBLP:journals/corr/abs-2405-16933,wei2025ainativememory20second}, which support behavior continuity and personalized memory modeling.
 
The third stage advances toward \textit{Systematic Memory Management}, integrating tool-based operations with OS-inspired governance frameworks. This includes toolkits such as EasyEdit and Mem0, which support explicit memory manipulation \cite{zhang_easyedit_2024,xu_easyedit2_2025,chhikara2025mem0buildingproductionreadyai}, as well as systems like Letta \cite{packer_memgpt_2024}, which implement paged context management and modular invocation. However, these systems still fall short of providing unified scheduling, lifecycle governance, and memory fusion across roles or agents.

\section{\textsc{MemOS} Design Philosophy}

As AGI continues to evolve into increasingly complex systems characterized by multi-tasking, multi-role collaboration, and multi-modality, language models must move beyond merely “understanding the world”—they must also “accumulate experience,” “retain memory,” and “continuously evolve.” However, prevailing architectures remain anchored in static parameters and lack structured modeling and unified management of memory, rendering them inadequate for supporting knowledge updates, state retention, and personalized adaptation. We propose that treating memory as a first-class resource and building a memory-centric execution paradigm is key to enabling continual adaptation and long-term reasoning in future LLMs.

\begin{wrapfigure}{r}{0.5\textwidth}
    \centering
    \vspace{-1.5em}
    \includegraphics[width=0.48\textwidth]{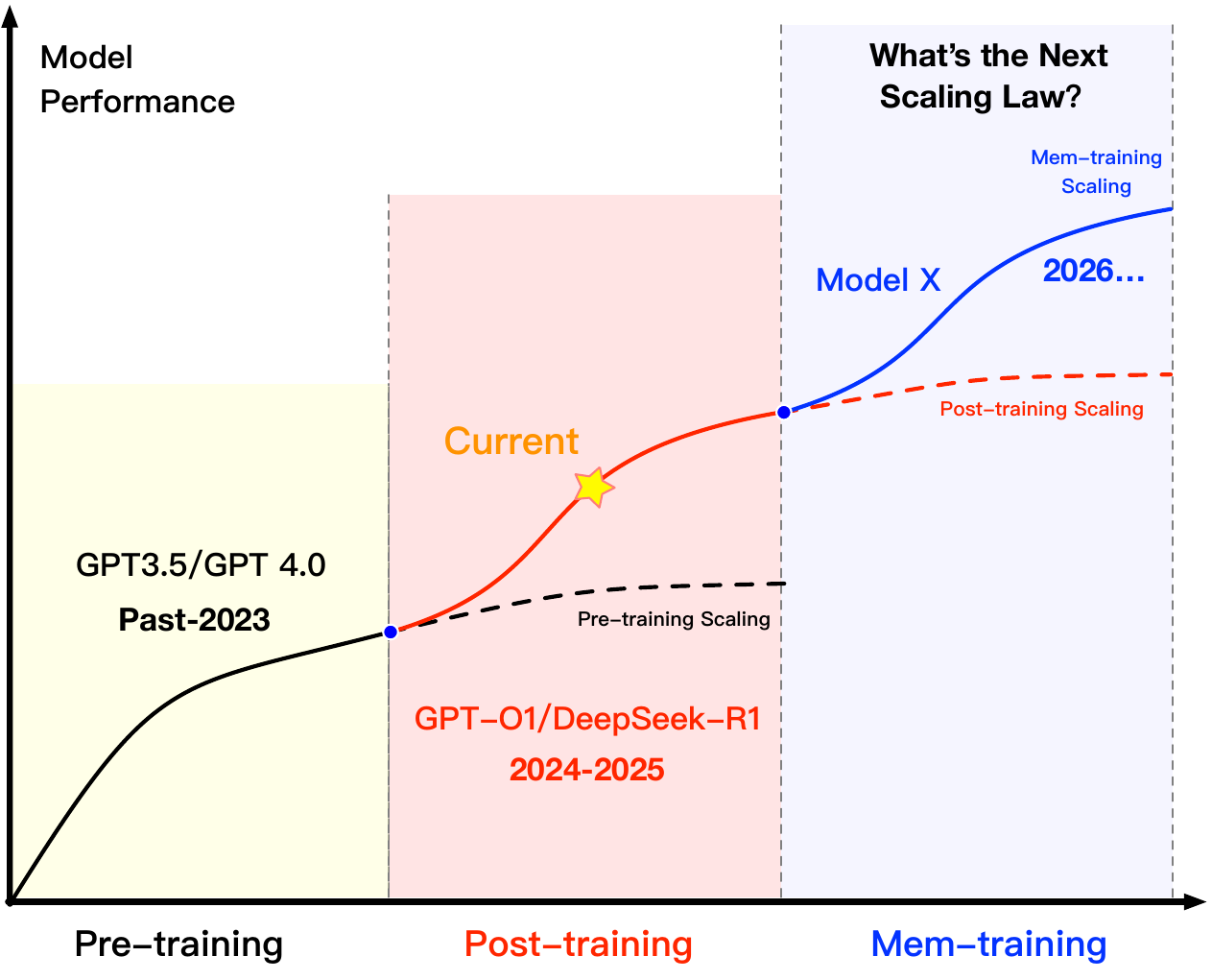}
    \caption{The next leap in model capability evolution hinges on the introduction of memory systems, marking a paradigm shift toward ``memory training''.}
    \label{fig:memos_scaling}
    \vspace{-1em}
\end{wrapfigure}

As shown in Figure~\ref{fig:memos_scaling}, traditional scaling laws are approaching diminishing returns. The research paradigm is shifting from data- and parameter-centric pretraining to post-training paradigms focused on alignment and fine-tuning. Yet even this refined approach faces dual challenges: diminishing performance gains and increasing engineering complexity. We posit that the next fundamental leap will arise from the ability to continuously model and schedule memory—enabling LLMs to maintain contextual consistency, adapt to evolving knowledge, and support iterative refinement across tasks.

To this end, we introduce \textsc{MemOS}—a prototype system designed to support a new \textit{memory-centric training paradigm}, where learning and inference are no longer separate phases but part of a unified, memory-driven process. \textsc{MemOS} not only enables structured memory storage, interface-level invocation, and lifecycle management, but also provides unified scheduling and version control mechanisms that constitute the foundational infrastructure for sustainable intelligence evolution. In our design vision, \textsc{MemOS} treats memory as a schedulable core resource, breaking down silos between agents, users, applications, and sessions. It adopts \textit{evolution} as a central management objective—supporting memory recomposition, migration, and fusion to facilitate long-term capability growth. Simultaneously, \textit{governance} is a foundational pillar: \textsc{MemOS} integrates access control, traceability, and interpretability mechanisms to ensure safe and compliant model operation in complex environments.

\section{\textsc{MemOS}}

\subsection{Types of Memory in \textsc{MemOS}}

In \textsc{MemOS}, memory is not merely a container of knowledge, but serves as the continuous substrate for perception, understanding, and action within the model. To systematically support LLMs in evolving across diverse tasks and scenarios, \textsc{MemOS} classifies memory into three core types: \textbf{Parametric Memory}, \textbf{Activation Memory}, and \textbf{Plaintext Memory}. Each type differs in its representation, lifecycle, and invocation mechanism, collectively forming the multi-layered structure of an intelligent agent’s cognitive system.

\noindent\textbf{Parametric Memory} refers to long-term knowledge encoded directly into model weights through pretraining or fine-tuning, embedded within feedforward and attention layers. It can participate in inference without the need for external retrieval. This memory type underpins fundamental language understanding, general knowledge, and skill modules—serving as the backbone for zero-shot generation and capability-driven agents. In \textsc{MemOS}, parametric memory includes not only foundational language capabilities but also supports modular, domain-specific injection—such as legal or medical knowledge—via pluggable LoRA-based modules for efficient composition and reuse.

\noindent\textbf{Activation Memory} denotes the transient cognitive state generated during inference, including hidden layer activations, attention weights, and KV-cache structures. It plays a critical role in context awareness, instruction alignment, and behavior modulation. \textsc{MemOS} treats activation memory as a “working memory” layer, enabling dynamic scheduling for tasks such as context persistence, stylistic control, and behavioral supervision. Frequently accessed activation states—such as KV-caches or attention patterns—can be transformed into semi-structured fragments or parametric modules, allowing short-term memory to persist and evolve over time.

\noindent\textbf{Plaintext Memory} comprises explicit knowledge retrieved from external sources, characterized by properties such as editability, shareability, and governance compatibility. Typical formats include documents, knowledge graphs, and prompt templates. This memory type addresses the limitations of context window size and fixed parameters, enabling rapid knowledge updates, personalized injection, and multi-agent collaboration. In \textsc{MemOS}, plaintext memory contributes to inference context generation and supports versioning, access control, and invocation tracing—serving as the foundation of knowledge governance.

These three types of memory are unified under a standard operational abstraction in \textsc{MemOS}: the \hyperlink{memcube}{\texttt{Memory Cube (MemCube)}}, which supports cross-type scheduling, lifecycle management, and structured fusion. By enabling transformation pathways between memory types (e.g., Activation $\rightarrow$ Plaintext, Plaintext $\rightarrow$ Parametric), \textsc{MemOS} establishes a scalable memory runtime that elevates LLMs from mere language generators to memory-enabled, adaptive, and continually evolving agents.

\begin{figure}[htp]
    \centering
    \includegraphics[width=0.8\linewidth]{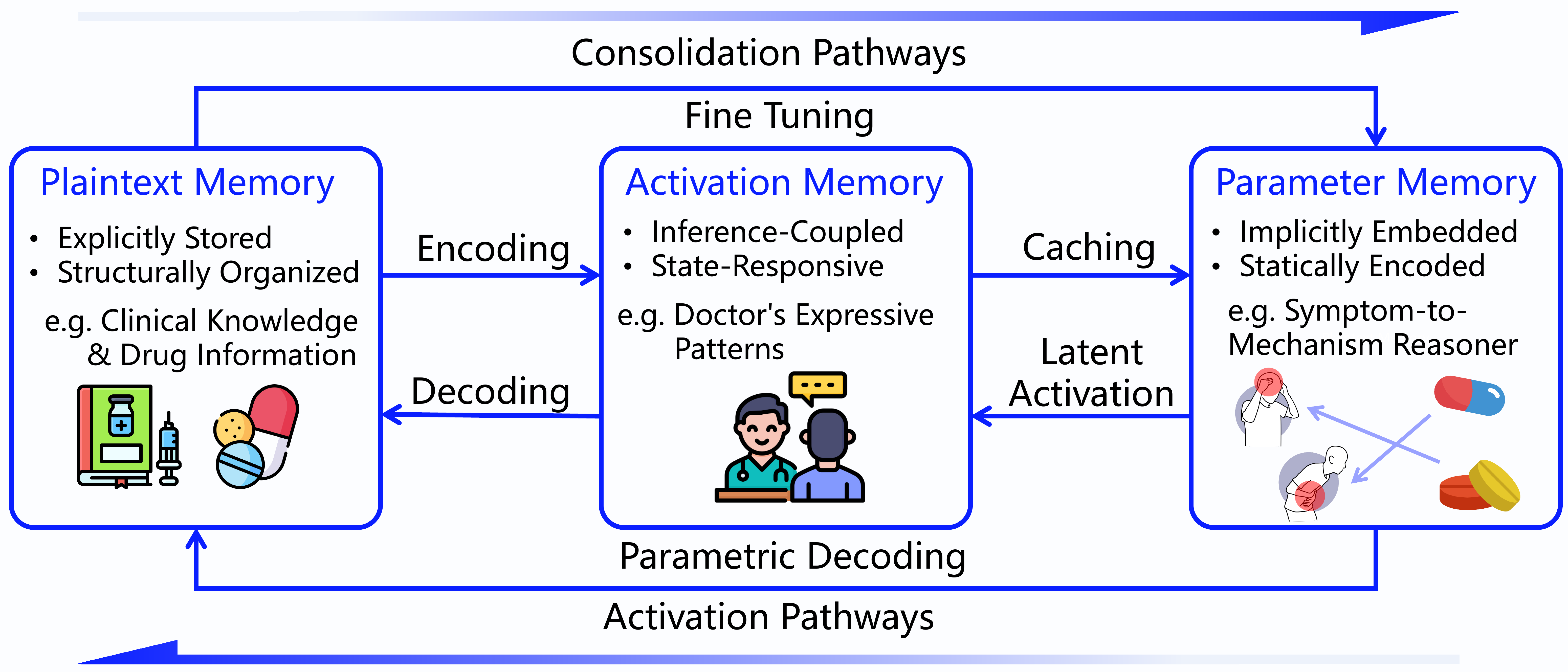}
    \caption{Transformation paths among three types of memory, forming a unified, controllable, and evolvable memory space.}
    \label{fig:memories}
\end{figure}

\subsection{Memory Cube (MemCube) as a Core Resource}
\label{sec:memcube}

In \textsc{MemOS}, the key to unifying and evolving heterogeneous memory resources lies in standardizing their representation and management mechanisms. To this end, we introduce \hypertarget{memcube}{\textbf{\texttt{MemCube}}} as the system’s fundamental encapsulation unit (see Figure~\ref{fig:MemCube}). The memory resources of LLMs span parametric knowledge, KV-caches, and externally injected content—each differing in origin, lifecycle, and invocation semantics. \texttt{MemCube} unifies these heterogeneous forms through a consistent data structure and interface, encapsulating both a semantic payload and structured metadata to enable uniform scheduling, access control, and lifecycle governance. \texttt{MemCube} metadata is organized into three categories to support memory identification, control, and evolution:

\begin{figure}[htp]
    \centering
    \includegraphics[width=0.8\linewidth]{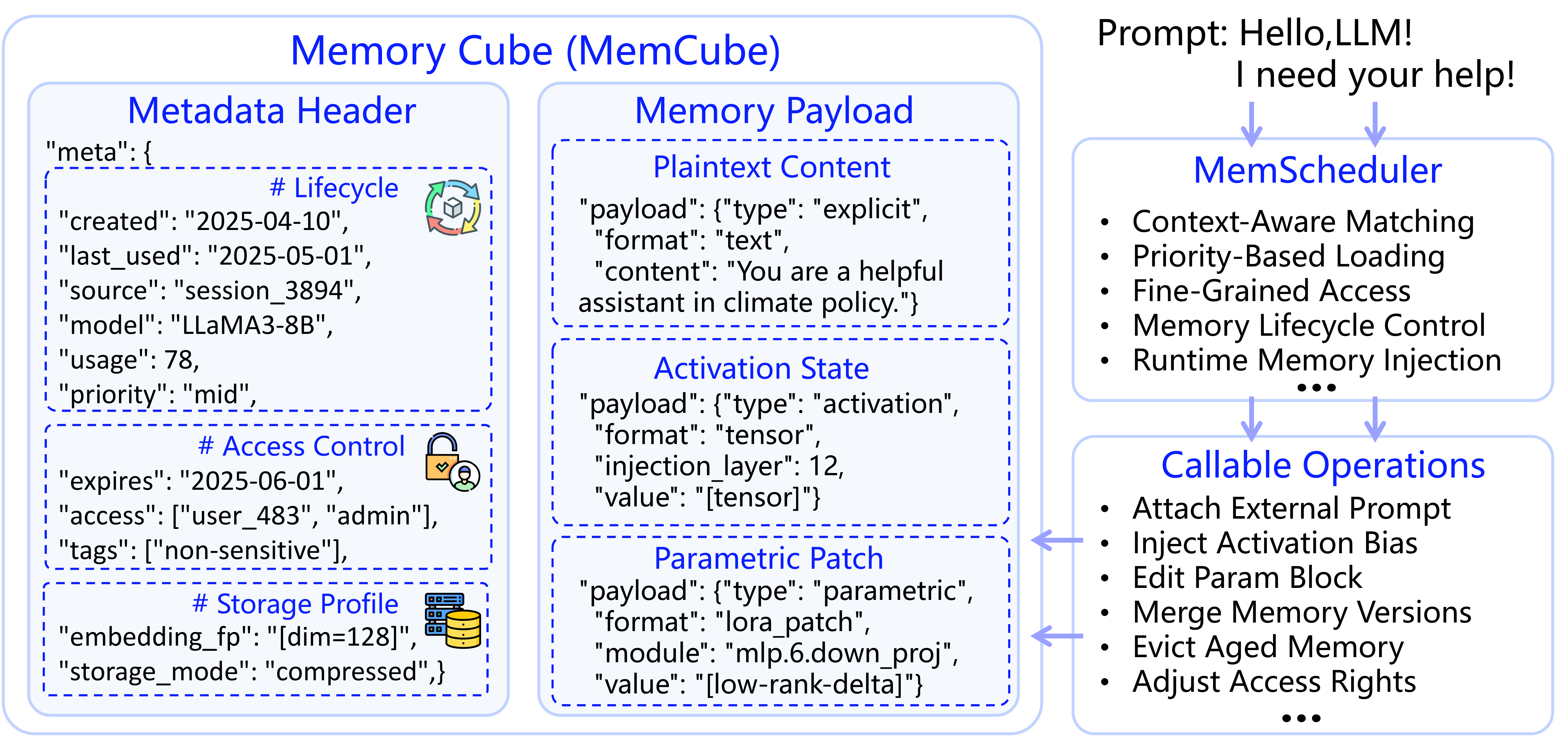}
    \caption{MemCube: a unified abstraction for heterogeneous memory, comprising a metadata header and semantic payload—serving as the smallest execution unit of memory in \textsc{MemOS}.}
    \label{fig:MemCube}
\end{figure}

\paragraph{\textbf{Descriptive Metadata}}
Used to identify the memory unit and define its semantic role. This includes timestamps (for creation or updates), origin signatures (e.g., user input, inference output), and semantic types (e.g., user preference, task prompt, domain knowledge).

\paragraph{\textbf{Governance Attributes}}
Enable safe and controlled usage in multi-user environments. These include access permissions, lifespan policies (e.g., time-to-live or frequency-based decay), priority levels, and compliance mechanisms such as sensitivity tags, watermarking, and access logging.

\paragraph{\textbf{Behavioral Indicators}}
Capture runtime usage patterns—automatically collected metrics such as access frequency, context relevance, and version lineage—that inform dynamic scheduling and cross-type transformation. This mechanism supports automatic adaptations, such as:

\begin{itemize}[noitemsep,leftmargin=1.5em]
    \item \textbf{Plaintext $\Rightarrow$ Activation}: Frequently accessed plaintext memory is converted into activation templates to reduce re-decoding costs;
    \item \textbf{Plaintext/Activation $\Rightarrow$ Parametric}: Stable, reusable knowledge is distilled into parametric structures to boost inference efficiency;
    \item \textbf{Parametric $\Rightarrow$ Plaintext}: Rarely used or outdated parameters are externalized into editable plaintext for greater flexibility.
\end{itemize}

With contextual fingerprinting and policy-aware scheduling, the system enables on-demand activation, hierarchical caching, and structural evolution—making \texttt{MemCube} a self-aware and continuously adaptive memory unit.

\subsection{\textsc{MemOS} Architecture} 

\begin{figure}[htp]
    \centering
    \includegraphics[width=1.0\linewidth]{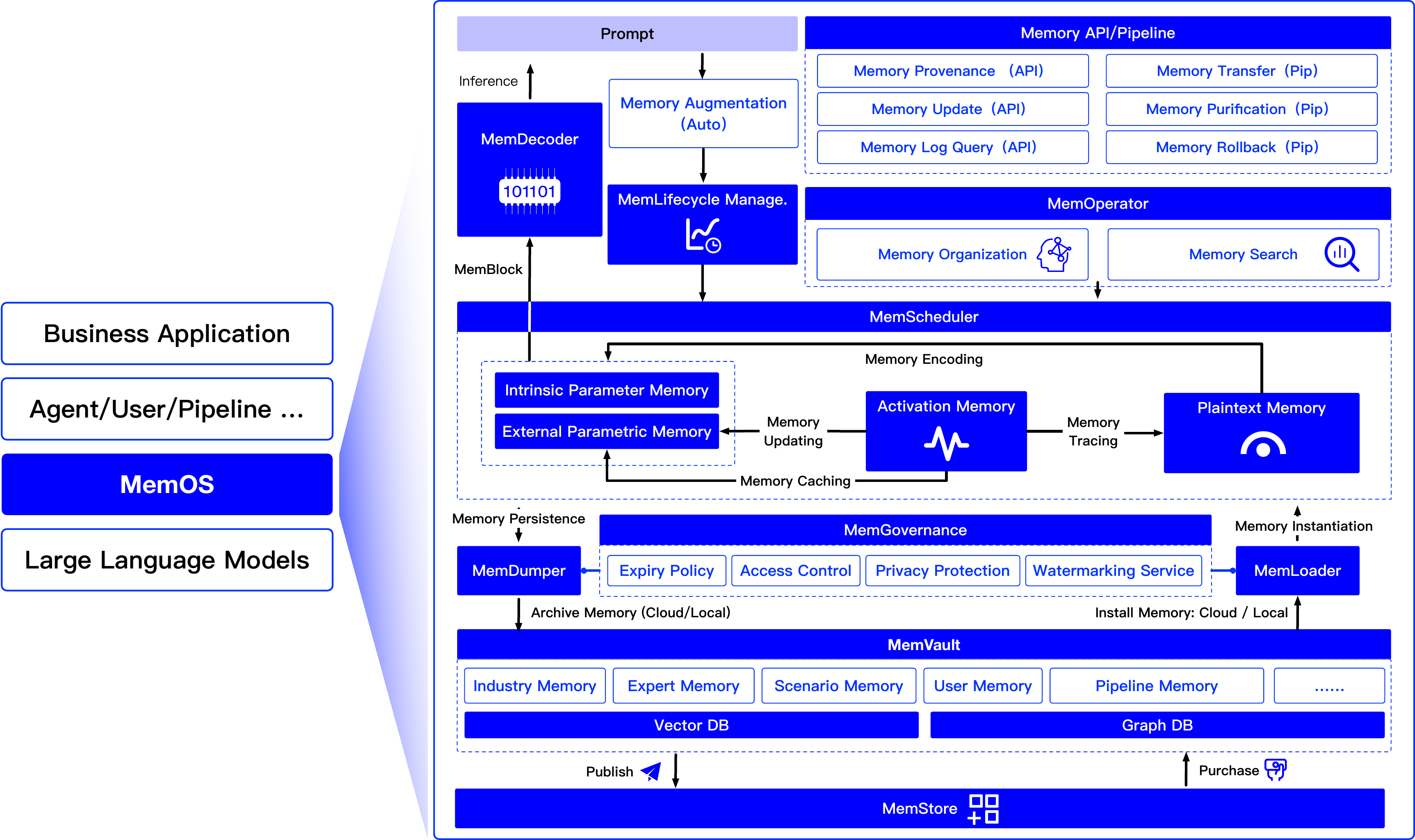}
    \caption{Overview of the \textsc{MemOS} architecture: showing the end-to-end memory lifecycle from user input to API parsing, scheduling, activation, governance, and evolution—unified via MemCube.}
    \label{fig:framework}
\end{figure}

To support unified and adaptive memory handling in LLMs,  provides an execution framework for memory parsing, scheduling, and governance. As shown in Figure~\ref{fig:framework}, it manages the full memory lifecycle via the \texttt{MemoryCube} abstraction. \textsc{MemOS} adopts a modular three-layer architecture, forming a closed-loop memory governance framework across the Interface Layer, Operation Layer, and Infrastructure Layer (see Figure~\ref{fig:overview}).

The \textbf{Interface Layer} serves as the system entry point, responsible for parsing natural language requests, identifying memory-related intents, and invoking standardized Memory APIs. The built-in \texttt{MemReader} component translates user inputs into structured memory operation chains.
The \textbf{Operation Layer} functions as the central controller, orchestrating components such as \texttt{MemScheduler}, \texttt{MemLifecycle}, and \texttt{MemOperator} to support task-aware scheduling, lifecycle control, and structural organization across users and workflows.
The \textbf{Infrastructure Layer} provides the foundational support for reliable execution, offering memory storage, access control, and cross-platform interoperability through modules such as \texttt{MemVault}, \texttt{MemGovernance}, and \texttt{MemStore}.

\paragraph{\textbf{Interface Layer: Memory API and Pipeline}}  
 
The Interface Layer is centered around a unified \texttt{Memory API}, offering key interfaces including \texttt{Provenance API}, \texttt{Update API}, and \texttt{LogQuery API}—used respectively for annotating memory sources, updating memory contents, and querying usage traces. All operations are encapsulated within the \texttt{MemoryCube} structure and governed by access control mechanisms provided through \texttt{MemGovernance}. To support multi-stage and composable workflows, \textsc{MemOS} introduces a pipeline-style operation chain mechanism. Each pipeline node transmits context, state, and intermediate outputs via \texttt{MemoryCube}, enabling transaction control, customizable topologies, and DAG-based scheduling. Developers can construct common operation patterns (e.g., Query–Update–Archive) to enable reuse across multi-model collaboration scenarios and ensure consistent memory operations.

\paragraph{\textbf{Operation Layer: Memory Scheduling and Lifecycle Management}}  
 
The Operation Layer orchestrates memory scheduling, lifecycle evolution, and organization. \texttt{MemScheduler} dynamically selects parametric, activation, or plaintext memory based on user-, task-, or organization-level context, supporting pluggable strategies such as least-recently-used (LRU), semantic similarity, and label-based matching. \texttt{MemLifecycle} models the memory lifecycle as a state machine and supports version rollback and freezing mechanisms to ensure auditability and temporal consistency. \texttt{MemOperator} manages memory through tagging systems, graph-based structures, and multi-layer partitions, enabling hybrid structural and semantic search. Retrieved results are linked back to \texttt{MemScheduler} to determine activation paths. Frequently accessed memory entries are cached at an intermediate layer to optimize performance. Collectively, these components enable effective structuring, precise invocation, and robust reasoning across tasks and agents.

\paragraph{\textbf{Infrastructure Layer: Governance and Memory Store}}  
 
The Infrastructure Layer governs memory compliance, storage, and circulation, ensuring system trustworthiness and long-term evolvability. \texttt{MemGovernance} enforces access permissions, lifecycle policies, and audit trails to ensure secure and accountable memory operations in multi-user environments. \texttt{MemVault} manages diverse memory repositories and provides unified access across heterogeneous storage backends. \texttt{MemLoader} and \texttt{MemDumper} facilitate structured memory migration across platforms and agents while preserving contextual integrity. \texttt{MemStore} supports the open publishing and subscription of memory units, enabling multi-model knowledge sharing and collaborative execution.

Overall, the system operates through a closed-loop Memory I/O Path, with all modules interfacing via the \texttt{MemoryCube} abstraction. It supports view customization, access isolation, and extensibility to future multi-modal scenarios.

\subsection{System Execution Flow}

\begin{figure}[htp]
    \centering
    \includegraphics[width=0.8\linewidth]{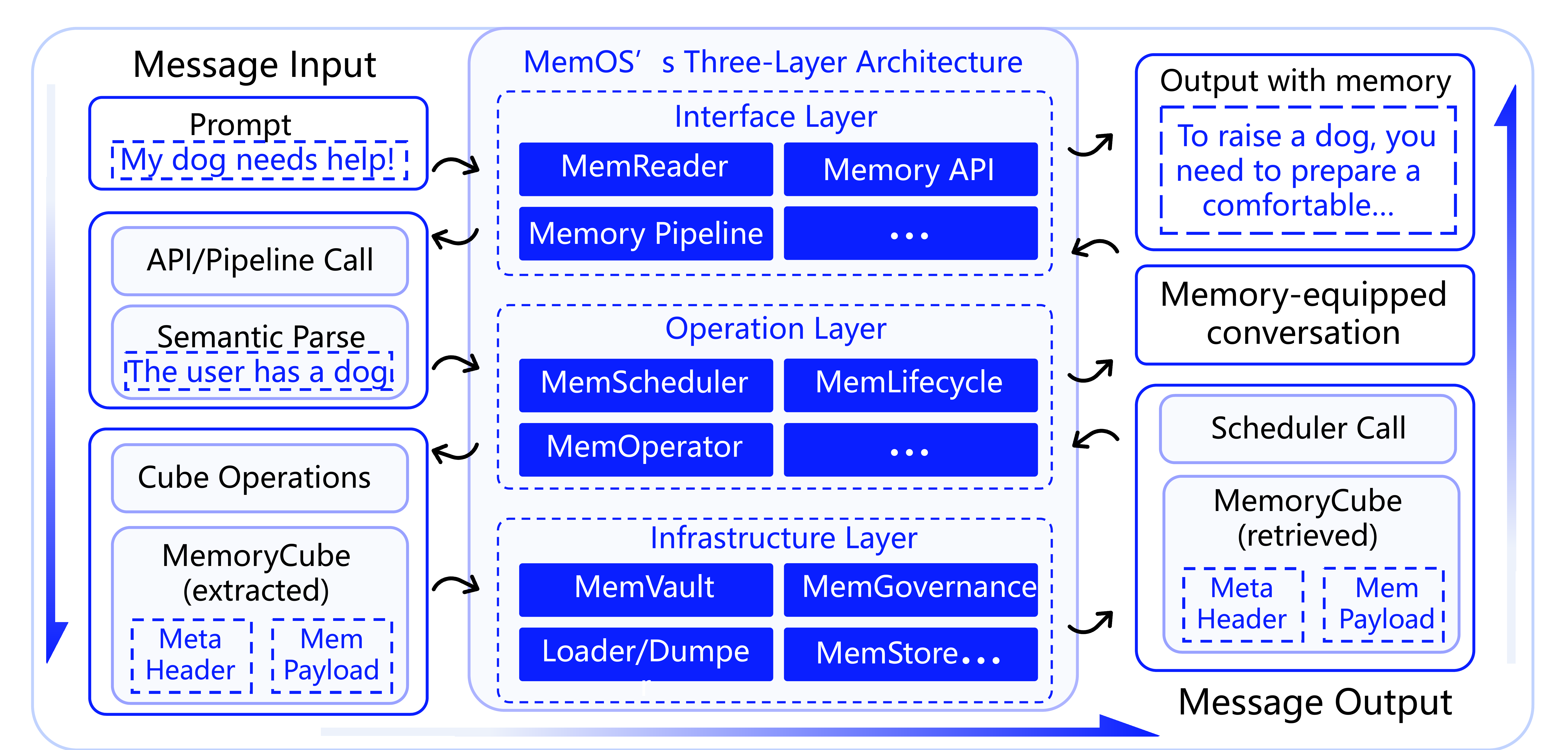}
    \caption{The three-layer architecture and memory I/O path of \textsc{MemOS}. From user input to scheduling and memory injection to response generation, each phase is executed via standardized \texttt{MemoryCube} structures that enable traceable and structured memory lifecycle management.}
    \label{fig:overview}
\end{figure}

As illustrated in Figure~\ref{fig:overview}, a \textsc{MemOS} execution begins with a user prompt or triggered task, parsed by \texttt{MemReader} into a structured \texttt{Memory API} call. This call initiates a pipeline, where context and state are passed via \texttt{MemoryCube} units.
\texttt{MemScheduler} then selects relevant memory (parametric, activation, or plaintext) based on access patterns and scheduling policies. Retrieved units are injected into the reasoning context. \texttt{MemOperator} organizes memory semantically and structurally, while \texttt{MemLifecycle} governs state transitions.
Archived memory is persisted in \texttt{MemVault}, managed by \texttt{MemGovernance}, and can be uploaded to or downloaded from \texttt{MemStore} for inter-agent collaboration. Migration between agents is supported by \texttt{MemLoader}/\texttt{MemDumper}.
This process forms a closed-loop memory flow—from input to activation, transformation, storage, and reuse—driven by declarative policies and executed through the \texttt{MemoryCube} abstraction.

\vspace{-1em}
\section{Conclusion}

In this work, we introduce a memory operating system designed for Large Language Models, aimed at collaboratively building foundational memory infrastructure for next-generation LLM applications.

\textsc{MemOS} provides a unified abstraction and integrated management framework for heterogeneous memory types, including parametric memory, activation memory, and explicit plaintext memory. We propose a standardized memory unit, \texttt{MemCube}, and implement key modules for scheduling, lifecycle management, structured storage, and transparent augmentation. These components collectively enhance reasoning coherence, adaptability, and system scalability in LLMs.
Building on this foundation, we envision a future intelligent ecosystem centered on modular memory resources and supported by a decentralized memory marketplace. This paradigm shift enables the creation of next-generation AI systems capable of continual learning and long-term evolution.

Looking ahead, we plan to explore the following directions:
\begin{itemize}
\item \textbf{Cross-LLM Memory Sharing}: Enable interoperability and module reuse across different foundation models by sharing parametric and activation memories. To support consistent semantics and secure exchange, we plan to extend the \textbf{Memory Interchange Protocol (MIP)} to define standard formats, compatibility rules, and trust mechanisms for cross-model/app memory transmission—facilitating collaborative knowledge transfer among agents.

\item \textbf{Self-Evolving MemBlocks}: Develop memory units capable of self-optimization, reconstruction, and evolution based on usage feedback, reducing the need for manual maintenance and supervision.
\item \textbf{Scalable Memory Marketplace}: Establish decentralized mechanisms for memory exchange, supporting asset-level transactions, collaborative updates, and distributed evolution to foster a sustainable AI ecosystem.
\end{itemize}

Overall, with the introduction of \textsc{MemOS}, we aim to transition LLMs from closed, static generation systems to continuously evolving intelligent agents equipped with long-term memory, integrated knowledge, and behavioral plasticity. \textsc{MemOS} not only addresses critical architectural limitations in current models but also lays the groundwork for cross-task, cross-platform, and multi-agent collaborative intelligence. We look forward to advancing the frontiers of \textsc{MemOS} in collaboration with the community, making memory a first-class computational resource in the age of general-purpose AI.

\bibliographystyle{plainnat}
\bibliography{main}

\end{document}